\documentclass[letterpaper]{article} 
\usepackage[submission]{aaai2027}  
\usepackage[hyphens]{url}  
\usepackage{graphicx} 
\usepackage{natbib}  
\usepackage{caption} 
\usepackage{subcaption}
\usepackage{algorithm}
\usepackage{algorithmic}
\usepackage{amsmath}
\usepackage{amssymb}
\usepackage{booktabs}
\usepackage{multirow}
\usepackage{adjustbox}
\usepackage{enumitem}
\setlist[itemize]{noitemsep, topsep=0pt}

\usepackage{newfloat}
\usepackage{listings}
\DeclareCaptionStyle{ruled}{labelfont=normalfont,labelsep=colon,strut=off} 
\floatstyle{ruled}
\newfloat{listing}{tb}{lst}{}
\floatname{listing}{Listing}

\usepackage{booktabs}

\title{PLAN: Parallel Liquid-Inspired Approximation Network for Efficient Representation Learning in Flexible Job Shop Scheduling}
\author {
    Dhivya Dharshini Kannan\textsuperscript{\rm 1},
    Wei Zhang\textsuperscript{\rm 1}\corresponding,
    Jieyi Bi\textsuperscript{\rm 2},
    Yingpeng Du\textsuperscript{\rm 2},
    Tianjun Wei\textsuperscript{\rm 2},
    Jie Zhang\textsuperscript{\rm 2},
    Zuming Liu\textsuperscript{\rm 3},
    Anupam Trivedi\textsuperscript{\rm 4}
}
\affiliations {
    \textsuperscript{\rm 1}Singapore Institute of Technology (SIT)\\
    \textsuperscript{\rm 2}Nanyang Technological University (NTU)\\
    \textsuperscript{\rm 3}Shanghai Jiao Tong University\\
    \textsuperscript{\rm 4}Agency for Science, Technology and Research (A*STAR)\\
    
}

\begin{document}

\maketitle

\begin{abstract}
Deep reinforcement learning (DRL) approaches for flexible job shop scheduling (FJSP) heavily rely on attention-centric architectures to achieve state-of-the-art performance. However, these models suffer from excessive parameter counts and prohibitive inference latency as problem scales expand. While liquid neural networks (LNNs) offer a parameter-efficient alternative for modeling adaptive state evolution, their inherently sequential dynamics bottleneck computational efficiency. To resolve this trade-off, we propose \textbf{PLAN} (Parallel Liquid-inspired Approximation Network), a lightweight representation learning framework that reformulates continuous liquid-state dynamics into a discretized and parallelizable formulation. PLAN structurally decouples state evolution from context aggregation, where liquid-inspired updates handle the primary evolving state representation, and a lightweight context aggregation module provides complementary global context. Furthermore, PLAN acts as a versatile, plug-and-play backbone that generalizes to complex FJSP variants, pairing with a compact stochastic module for stochastic FJSP and replacing heavy heterogeneous graph transformers in multi-faceted dynamic FJSP. Extensive evaluations across deterministic, stochastic, and multi-faceted dynamic FJSP benchmarks show that PLAN reduces the average makespan by 1.2\%, 1.4\%, and 2.3\%, respectively, compared with the corresponding state-of-the-art baselines, with the improvement reaching 10.2\% in one benchmark setting. PLAN also reduces average inference latency by 13.2\%, 31.7\%, and 26.9\%, respectively, with a maximum reduction of 69.2\% on the largest instances, while using only 22$-$47\% of the baseline parameters.
\end{abstract}


\section{Introduction}
Job scheduling is a fundamental combinatorial optimization problem with broad applications across industrial, computing, and service systems \cite{kwan2026rela}. Flexible Job Shop Scheduling (FJSP), one of its most widely studied formulations, has been applied to areas such as automotive assembly \cite{ADP}, healthcare scheduling \cite{ATA}, and semiconductor fabrication \cite{FMS}. In FJSP, each operation of a job can be assigned to one of multiple eligible machines while satisfying precedence and resource constraints \cite{MLD}. Due to its NP-hard nature \cite{ROF}, obtaining high-quality schedules within acceptable computation time becomes increasingly difficult as the problem size grows. Classical methods, including tabu search \cite{mkdir}, genetic algorithms \cite{Hurink}, and dispatching heuristics \cite{AEH}, often struggle to achieve a good trade-off between solution quality and computational efficiency, especially in large-scale or dynamic scheduling environments.

Recent progress in deep reinforcement learning (DRL) has demonstrated the potential of learning-based schedulers to generate high-quality schedules with fast inference after training \cite{AND,FJS}. Existing state-of-the-art (SOTA) DRL-based approaches, such as HGNN \cite{HGNN} and DANIEL \cite{DANIEL}, achieve strong performance through deep attention-based representation learning. However, these architectures rely on multiple full attention blocks and scheduling-specific auxiliary components to model interactions among operations and machines, increasing parameter count, memory footprint, and inference latency. As the numbers of operations and machines grow, these components must capture increasingly complex interactions, resulting in higher computational overhead for large scheduling instances.

Scheduling decisions are made sequentially, and each decision immediately changes the scheduling state, including machine availability and operation readiness. The scheduling state therefore evolves throughout the decision process. Deep attention-based architectures are effective at modelling interactions among scheduling entities \cite{HGNN,DANIEL}. However, they are not explicitly designed to model the decision-dependent evolution of scheduling states throughout the sequential decision process. This motivates the exploration of alternative representation learners that can propagate scheduling information efficiently while remaining compact. Liquid neural networks (LNNs), originally designed for continuous-time dynamic systems, support adaptive state updates with few parameters \cite{LNN}, making them a promising mechanism for representing evolving states. However, their inherently sequential state evolution limits computational efficiency and prevents the state updates from being efficiently processed in parallel.

Motivated by these observations, we propose PLAN, a Parallel Liquid-Inspired Approximation Network, as a lightweight representation learning framework for DRL-based FJSP. PLAN reformulates the sequential liquid-state evolution of LNNs into a discretized and parallelizable representation learning process. Its liquid-inspired state updates perform the primary representation learning, while a shallow attention module provides complementary global context. This design shifts the main representation learning from deep attention to liquid-inspired state updates, substantially reducing architectural complexity. Experiments under deterministic, stochastic, and multi-faceted dynamic FJSP settings show that PLAN reduces model complexity and inference latency while improving scheduling performance. 

The main contributions are summarized as follows.

\begin{itemize}
    \item We propose PLAN, a lightweight representation learner combining liquid-inspired state updates with a shallow attention module for efficient FJSP scheduling.
    
    \item We develop a parallelizable liquid-inspired representation learner by reformulating the sequential ordinary differential equation (ODE) dynamics of LNNs through an Euler-based approximation, preserving adaptive state updates while enabling parallel computation.
    
    \item We extend PLAN to stochastic FJSP using a smaller stochastic processing module (SPM) and to multi-faceted dynamic FJSP by replacing the original heterogeneous graph transformer (HGT). Experiments across deterministic and dynamic settings show reduced model size and inference latency together with improved performance.
\end{itemize}

\section{Problem Formulation and Scheduling Settings}
FJSP includes sets of jobs $J={J_1,J_2,\ldots,J_n}$ and machines $M={M_1,M_2,\ldots,M_m}$, where each job $J_i$ consists of an ordered sequence of operations $O_i={O_{i1},O_{i2},\ldots,O_{in_i}}$, and $n_i$ denotes the number of operations in $J_i$. The full set of operations is denoted as $O=\bigcup_i O_i$. An operation $O_{ij}$ is assigned to one machine from its compatible machine set $M_{ij}\subseteq M$. When $O_{ij}$ is processed on machine $M_k\in M_{ij}$, it requires processing time $p_{ij}^{k}>0$, and $C_{ij}$ denotes its completion time. The objective is to minimize the makespan $C_{\max}$, i.e., the completion time of the last completed operation.
\begin{equation}
C_{\max}=\max_{O_{ij}\in O} C_{ij},
\end{equation}
A feasible schedule must satisfy the precedence constraints within each job, assign exactly one compatible machine to each operation, and ensure that each machine processes at most one operation at a time.

Processing times may be uncertain due to factors such as resource conditions, execution delays, and unexpected disturbances, and their exact values may be unavailable before scheduling. To model this uncertainty, we consider stochastic FJSP with stochastic processing times~\cite{NCO}, where the deterministic processing time $p_{ij}^{k}$ is replaced by a random variable $P_{ij}^{k}$, making the operation completion times and final makespan random variables. We further evaluate PLAN under the multi-faceted dynamic FJSP setting, following the benchmark configuration and dynamic-event protocol established in~\cite{MFDFJSP}.



\begin{figure*}
    \centering
    \includegraphics[width=0.95\linewidth]{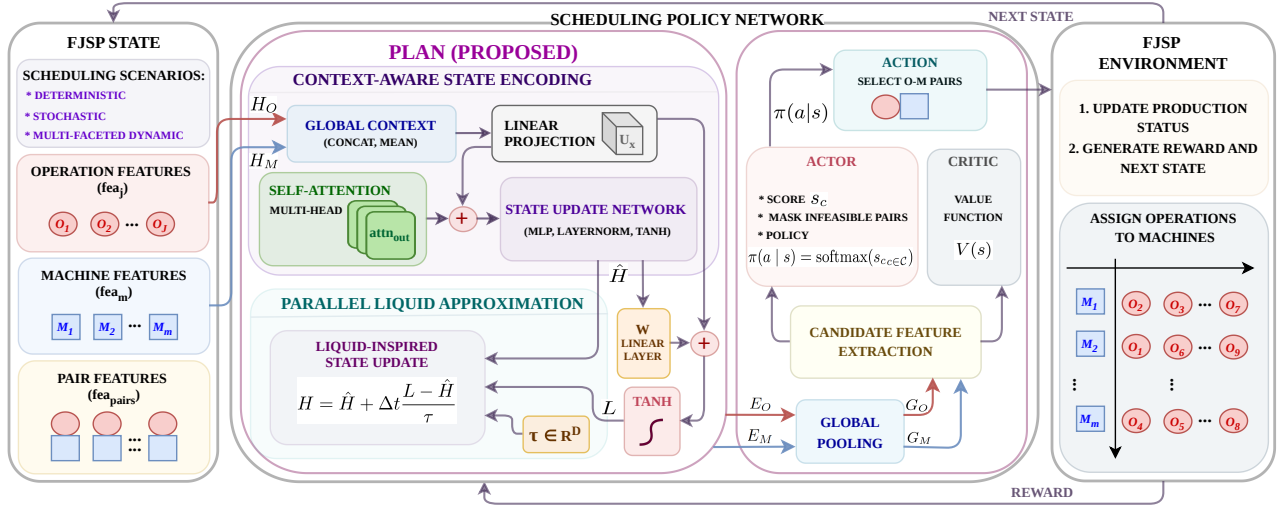}
    \caption{Overview of the proposed PLAN framework for FJSP.}
    \label{fig:PLAN}
\end{figure*}

\section{Methodology}
This section presents the DRL formulation, the PLAN framework and its key components and the learning procedure.

\subsection{MDP Formulation}
The scheduling problem is formulated as a Markov decision process (MDP), where operation-machine assignment decisions are made sequentially until all operations have been assigned to machines. At each decision step, a DRL agent selects an operation-machine pair based on the current scheduling state and receives a reward reflecting the quality of the resulting schedule~\cite{DANIEL}. The MDP is defined by the state space $\mathcal{S}$, action space $\mathcal{A}$, transition function $\mathcal{P}$, and reward function $\mathcal{R}$, which are described below.

\textbf{State}.
The state $s_t$ denotes the current scheduling status at decision step $t$. For deterministic FJSP, it consists of three categories of entity features, given by $s_t=\{H_O,H_M,H_{OM}\}$, where $H_O$, $H_M$, and $H_{OM}$ denote the operation, machine, and operation-machine pair features, respectively. The operation features describe the processing and scheduling status of operations, the machine features characterize machine utilization and availability, and the pair features capture the compatibility and processing relationships between candidate operations and machines. For stochastic FJSP, the state additionally includes sampled processing-time scenarios representing multiple possible realizations of processing-time uncertainty. For the multi-faceted dynamic setting, it further captures changes in the scheduling environment, such as machine breakdowns and new job arrivals.

\textbf{Action}.
At each decision step $t$, the agent selects a feasible action $a_t=(O_{ij},M_k)$ that assigns operation $O_{ij}$ to machine $M_k$. The action space $\mathcal{A}_t$ contains all feasible operation-machine pairs at decision step $t$ satisfying operation precedence and machine compatibility constraints. The same action definition is used across the deterministic, stochastic, and multi-faceted dynamic settings.

\textbf{State Transition}.
Once an action $a_t$ is executed, the scheduling environment updates the operation status, machine availability, and feasible action space according to the operation precedence and machine constraints, producing the next state $s_{t+1}$ from the current state $s_t$.

\textbf{Reward}.
The reward function is designed to encourage schedules with smaller makespan. At state $s_t$, the estimated makespan is denoted by $\hat{C}_{\max}(s_t)$. After an action, the immediate reward is formulated as the difference between the estimated makespan of the current and next states, $r_t=\hat{C}_{\max}(s_t)-\hat{C}_{\max}(s_{t+1})$. A positive reward indicates an improvement, while a negative reward indicates a reduction in scheduling quality.
For stochastic FJSP, the estimated makespan is evaluated over $n$ sampled processing-time scenarios, denoted by $\{\hat{C}_{\max}^{1}(s_t),\hat{C}_{\max}^{2}(s_t),\ldots,\hat{C}_{\max}^{n}(s_t)\}$. We adopt Value-at-Risk (VaR) as the risk-sensitive scheduling objective, i.e., $f(s_t)=\mathrm{VaR}_{\alpha}\big(\hat{C}_{\max}(s_t)\big)$, and define the immediate reward as $r_t=f(s_t)-f(s_{t+1})$.


\textbf{Policy}.
The policy $\pi_{\theta}(a_t|s_t)$ maps the current scheduling state to a probability distribution over feasible actions. The policy parameters $\theta$ are learned through interactions with the scheduling environment.

\subsection{PLAN Framework}
Figure~\ref{fig:PLAN} illustrates the overall architecture of PLAN. First, liquid-inspired state dynamics enable adaptive representation learning for the evolving scheduling environment. Second, parallel approximation enables efficient liquid state updates without sequential ODE integration, making the framework suitable for large-scale scheduling. To realize these ideas, PLAN aggregates global contextual information, and performs parallel liquid state updates to learn operation and machine representations for downstream scheduling decisions. Each component is described in detail below.

\subsubsection{Liquid-Inspired State Dynamics}
FJSP is a dynamic decision-making problem in which machine workloads, operation statuses, and feasible actions continuously evolve throughout the scheduling process. Therefore, the representation learning model should capture not only the relationships among scheduling entities but also the evolution of the scheduling state.
LNNs \cite{DLNET} naturally model such evolving states through adaptive state dynamics. Instead of learning a static mapping, the hidden state evolves continuously according to the current hidden state and scheduling input, allowing the representation to adapt as the scheduling environment changes. This formulation motivates PLAN, which develops an efficient parallel approximation for scheduling.
For input $x(t)$ and hidden state $h(t)$, the continuous liquid dynamics are formulated as,
\begin{equation}
\frac{\mathrm{d}h_t}{\mathrm{d}t} =
-\frac{h_t}{\tau}
+ \sigma\left(
W_h h_t + W_x x_t
\right),
\label{eq:dhdt}
\end{equation}
where $\tau$ is the learnable time constant, $W_h$ and $W_x$ are trainable weight matrices, and $\sigma(\cdot)$ denotes a nonlinear activation function. Given an initial hidden state $h_0$, the ODE in Eq.~(\ref{eq:dhdt}) is integrated over the time interval $[0,T]$ to obtain the evolved hidden state $h(T)$, where $T$ denotes the integration horizon,
\begin{equation}
h(T) = \mathrm{ODE}
\left(\frac{\mathrm{d}h}{\mathrm{d}t}, h_0
\right).
\label{eq:hT}
\end{equation}
Direct numerical integration introduces additional computational overhead and is not well suited to FJSP, where decisions are made at discrete scheduling steps. We therefore adopt a first-order Euler discretization with time step $\Delta t$,
\begin{equation}
h_{t+1} =
h_t + \Delta t \frac{\mathrm{d}h_t}{\mathrm{d}t}.
\label{eq:euler}
\end{equation}
Substituting Eq.~(\ref{eq:dhdt}) into Eq.~(\ref{eq:euler}) gives,
\begin{equation}
h_{t+1} =
h_t + \Delta t
\left(
-\frac{h_t}{\tau}
+ \sigma\left(
W_h h_t + W_x x_t
\right)\right).
\label{eq:ht-euler}
\end{equation}
Although Eq.~(\ref{eq:ht-euler}) converts the continuous dynamics into a discrete formulation, the hidden state is still updated sequentially because each state depends on the previously computed state. This sequential dependency limits parallel execution and reduces computational efficiency on modern hardware. To address this limitation, PLAN reformulates the liquid state update as a parallel approximation, as described below, while preserving the adaptive characteristics of liquid dynamics.

\subsubsection{Context-Aware Representation Learning}
The input state consists of heterogeneous operation and machine features. To enable unified representation learning, PLAN first projects each feature set $X$ into a common latent space through a trainable linear transformation $\mathcal{U}(\cdot)$, yielding $U_X=\mathcal{U}(X)$. Although the projected features preserve the local information of individual scheduling entities, they do not explicitly capture the contextual relationships among them. Therefore, PLAN employs a lightweight multi-head attention (MHA) module to aggregate global scheduling context. 
Unlike current SOTA schedulers that rely on deep attention blocks \cite{Transformer} as the main representation learner, PLAN uses attention only to aggregate scheduling context, while the liquid-inspired state update performs the main representation learning. Therefore, PLAN employs only a lightweight MHA module, without the stacked feed-forward, normalization, and residual blocks of a full Transformer encoder.
The contextual representation is computed as $A=\mathrm{MHA}(U_X,U_X,U_X)$, where the projected features serve as the query, key, and value to perform self-attention within the same feature set, enabling parallel information exchange and aggregation of global scheduling context. The projected features and contextual representation are then fused to estimate the initial hidden state,
\begin{equation}
Z=[U_X;A], \qquad
\widehat{H}=\phi(Z),
\label{eq:hidden}
\end{equation}
where $[\cdot;\cdot]$ denotes feature concatenation, and $\phi(\cdot)$ is the state estimation network consisting of two fully connected layers with layer normalization and a nonlinear activation. It maps the fused local and global contextual features to the initial hidden state for the subsequent liquid approximation.

\subsubsection{Parallel Liquid Approximation}
The estimated hidden states initialize the liquid dynamics. However, directly applying the Euler update in Eq.~(\ref{eq:ht-euler}) still requires recurrent state propagation, preventing all hidden states from being computed simultaneously. PLAN therefore approximates the liquid state evolution through a parallel formulation that preserves the adaptive characteristics of liquid dynamics while eliminating sequential dependencies. 
The liquid-inspired nonlinear response is first computed from the estimated hidden state as,
\begin{equation}
L=\tanh
\left(
W\hat{H}+U_X\right),
\end{equation}
where $W$ is a trainable weight matrix, and $U_X$ provides the projected scheduling features as the external input. The $\tanh$ activation preserves the bounded nonlinear state transition of the original liquid dynamics while operating on the estimated hidden state. The liquid state is then refined using a single liquid-inspired correction step,
\begin{equation}
H=\hat{H}+\Delta t
\frac{L-\hat{H}}
{\tau},
\label{eq:plan_update}
\end{equation}
which can be viewed as a parallel approximation of a single Euler-style liquid update without recurrent state propagation. Consequently, all hidden states can be refined simultaneously through batched matrix operations. The resulting representations capture both the contextual relationships among scheduling entities and the nonlinear state adaptation inherited from the liquid formulation, providing informative embeddings for downstream scheduling decisions.

\begin{algorithm}[t]
\caption{PLAN Encoding and Decision Procedure}
\label{alg:plan}
\small
\begin{algorithmic}[1]

\STATE \textbf{Input:} Operation features $H_O$, machine features $H_M$, candidate operations $\mathcal{C}$, and pair features $H_{OM}$

\STATE Encode operation context:
$H_O^{c}=[H_O;\operatorname{mean}(H_O)]$

\STATE Encode machine context:
$H_M^{c}=[H_M;\operatorname{mean}(H_M)]$

\FOR{$X \in {H_O,H_M}$}
\STATE Project features:
$U_X=\mathcal{U}(X)$
\STATE Aggregate scheduling context:
$A=\mathrm{MHA}(U_X,U_X,U_X)$
\STATE Fuse local and contextual features:
$Z=[U_X;A]$
\STATE Estimate hidden state:
$\hat{H}=\phi(Z)$
\STATE Compute liquid response:
$L=\tanh(W\hat{H}+U_X)$
\STATE Apply liquid approximation:
$H=\hat{H}+\Delta t(L-\hat{H})/\tau$
\ENDFOR

\STATE Obtain embeddings:
$G_O=\operatorname{pool}(H_O)$,
$G_M=\operatorname{pool}(H_M)$

\FOR{$c \in \mathcal{C}$}
\STATE Fuse representation:
$F_c=[H_O^{c};H_M^{c};G_O;G_M;H_{OM}^{c}]$
\STATE Compute candidate score:
$s_c=\mathrm{Actor}(F_c)$
\ENDFOR

\STATE Mask infeasible operation-machine pairs
\STATE Compute scheduling policy:
$\pi(a\mid s)=\operatorname{softmax}({s_c}_{c\in\mathcal{C}})$
\STATE Estimate state value:
$V(s)=\mathrm{Critic}([G_O;G_M])$

\STATE \textbf{Output:} Policy distribution $\pi(a\mid s)$ and state value $V(s)$
\end{algorithmic}
\end{algorithm}

\subsubsection{Operation and Machine Encoding}
PLAN processes the operation and machine features using two independent encoders with the same architecture. 
Before encoding, PLAN summarizes each feature set through mean pooling to capture its global scheduling context and concatenates this context with every corresponding entity feature, yielding $H_O^{c}=[H_O;\operatorname{mean}(H_O)]$ and $H_M^{c}=[H_M;\operatorname{mean}(H_M)]$, where $\operatorname{mean}(\cdot)$ computes the average feature vector across all entities and $[\cdot;\cdot]$ denotes feature concatenation.
The resulting context-enhanced features are then independently processed through the context-aware representation learning and parallel liquid approximation introduced above, producing the encoded representations $E_O$ and $E_M$, respectively. Finally, the encoded representations are pooled as $G_O=\operatorname{pool}(E_O)$ and $G_M=\operatorname{pool}(E_M)$ to obtain global operation and machine embeddings for the subsequent decision network.

\subsection{SPM-PLAN}
For the stochastic FJSP, PLAN is integrated with SPM \cite{NCO}, which summarizes sampled processing-time scenarios into a compact stochastic representation. For a set of $n$ scenario embeddings $H={h_1,h_2,\ldots,h_n}$, SPM avoids applying full self-attention across all scenarios. Instead, it employs a small set of inducing vectors $I$ to approximate their global interactions through two cross-attention blocks (CABs). The resulting stochastic representation is then obtained through mean pooling and formulated as,
\begin{equation}
\mathrm{SPM}(H)=
\operatorname{mean}\left(
\mathrm{CAB}
\bigl(
H,\mathrm{CAB}(I,H)
\bigr)\right).
\end{equation}
The stochastic representation is concatenated with the deterministic representation as $h=[h^{\mathrm{det}};\mathrm{SPM}(H)]$, enriching the scheduling state with processing-time uncertainty. Since PLAN already performs contextual representation learning and adaptive state refinement, a compact SPM is sufficient to extract stochastic information without introducing unnecessary computational overhead. Consequently, SPM-PLAN preserves effective uncertainty modeling while remaining lightweight and enabling faster inference.

\begin{table*}[!htbp]
\centering
\caption{Performance comparison on the small deterministic FJSP benchmarks SD1 and SD2. Gap (\%) denotes the relative makespan difference from the OR-Tools reference, and the average schedule generation time is reported. Lower values are better.}
\label{tab:small}
\begin{adjustbox}{width=\textwidth}
\begin{tabular}{c|c|c|ccc|ccc|ccc|ccc}
\toprule
\multirow{4}{*}{Data} &
\multirow{4}{*}{Size} &
\multirow{2}{*}{OR-Tools} &
\multicolumn{6}{c|}{Greedy} &
\multicolumn{6}{c}{Sampling} \\

\cmidrule(lr){4-15}

&&&
\multicolumn{3}{c|}{DANIEL} &
\multicolumn{3}{c|}{PLAN} &
\multicolumn{3}{c|}{DANIEL} &
\multicolumn{3}{c}{PLAN} \\

\cmidrule(lr){3-15}

&&
Makespan &
Makespan & Gap & Time &
Makespan & Gap & Time &
Makespan & Gap & Time &
Makespan & Gap & Time \\
\midrule

\multirow{4}{*}{SD1}
&10$\times$5
&96.32
&107.97& 12.10&0.48
&\textbf{107.36}&\textbf{11.47}&\textbf{0.33}
&102.36& 6.27&1.15
&\textbf{101.48}&\textbf{5.36}&\textbf{0.81}\\

&20$\times$5
&188.15
&\textbf{197.66}&\textbf{5.06}&0.90
&197.78&5.12&\textbf{0.65}
&193.83& 3.02&2.41
&\textbf{193.14}&\textbf{2.65}&\textbf{1.75}\\

&15$\times$10
&143.53
&160.78& 12.02&1.23
&\textbf{159.21}&\textbf{10.93}&\textbf{0.96}
&152.51& 6.26&3.83
&\textbf{150.68}&\textbf{4.98}&\textbf{3.04}\\

&20$\times$10
&195.98
&199.14& 1.61&1.62
&\textbf{198.21}&\textbf{1.14}&\textbf{1.29}
&195.30& $-$0.35&5.31
&\textbf{193.38}&\textbf{$-$1.33}&\textbf{4.30}\\

\midrule

\multirow{4}{*}{SD2}
&10$\times$5
&326.24
&413.91& 26.87&0.40
&\textbf{407.86}&\textbf{25.02}&\textbf{0.31}
&366.59& 12.37&1.11
&\textbf{362.46}&\textbf{11.10}&\textbf{0.81}\\

&20$\times$5
&602.04
&673.28& 11.83&1.03
&\textbf{661.45}&\textbf{9.87}&\textbf{0.83}
&632.85& 5.12&3.09
&\textbf{624.68}&\textbf{3.76}&\textbf{2.68}\\

&15$\times$10
&377.17
&588.14& 55.94&1.53
&\textbf{587.84}&\textbf{55.86}&\textbf{1.22}
&519.78& 37.81&4.93
&\textbf{515.61}&\textbf{36.70}&\textbf{4.36}\\

&20$\times$10
&464.16
&606.14& 30.59&2.01
&\textbf{603.43}&\textbf{30.00}&\textbf{1.65}
&552.09& 18.94&6.09
&\textbf{550.11}&\textbf{18.52}&\textbf{5.98}\\

\bottomrule
\end{tabular}
\end{adjustbox}
\end{table*}

\subsection{Training via PPO}
We train PLAN using proximal policy optimization (PPO). The actor network parameterized by $\theta$ produces the previously defined policy $\pi_{\theta}(a_t|s_t)$ over feasible actions. During training, actions are sampled from this distribution to encourage exploration, while interactions with the environment generate rewards and subsequent states to form scheduling trajectories. The critic network estimates the state value $V_{\phi}(s_t)$, which is used to compute advantage estimates that quantify the relative quality of sampled actions. PPO optimizes the actor through a clipped surrogate objective that limits excessive policy changes and stabilizes training. Through repeated interactions with the environment, the policy progressively learns to minimize the FJSP makespan.



\section{Experiments}
This section evaluates PLAN under deterministic, stochastic, and multi-faceted dynamic FJSP settings. 

\subsection{Datasets and Configuration}
We evaluate PLAN on deterministic, stochastic, and multi-faceted dynamic FJSP benchmarks. The deterministic evaluation uses the synthetic SD1 and SD2 datasets~\cite{DANIEL}, covering small ($10\times5$, $20\times5$, $15\times10$, $20\times10$), medium ($30\times10$, $40\times10$), and large ($100\times10$, $200\times5$) problem scales, together with the public Brandimarte~\cite{mkdir} and Hurink~\cite{Hurink} benchmarks. For stochastic scheduling, we adopt the SD3 benchmark~\cite{NCO}, where processing times are generated by sampling around the median deterministic processing times. For multi-faceted dynamic scheduling, we use the benchmark and DRL framework from~\cite{MFDFJSP}.
Each benchmark contains 100 instances for every problem scale. We evaluate both greedy and sampling action-selection strategies, where greedy selects the action with the highest policy probability and sampling draws actions from the policy distribution. All experiments are repeated with five random seeds (0–4), and the reported results are averaged across runs.

For deterministic FJSP, we adopt the implementation settings of DANIEL, the current SOTA, to ensure a fair comparison and compare PLAN against it. We additionally report OR-Tools as a reference solver. Since exact optimization becomes computationally expensive for large-scale FJSP with complex constraints, following standard practice, OR-Tools is executed with a 30-minute time limit for each instance, whereas DRL methods, e.g., PLAN, generate schedules almost instantly after training. For stochastic FJSP, we evaluate SPM-PLAN by integrating PLAN with SPM and compare it with SPM-DAN, the stochastic extension of DANIEL. For multi-faceted dynamic FJSP, we follow the implementation settings and evaluation protocol of HGT, the SOTA method for this setting. All models are implemented in PyTorch and trained on a workstation equipped with an NVIDIA RTX PRO 5000 Blackwell GPU with 48 GB of memory.

\subsection{Deterministic FJSP}
We compare PLAN with DANIEL and OR-Tools on the small deterministic benchmarks from SD1 and SD2, where training and testing use the same instance sizes. As shown in Table~\ref{tab:small}, PLAN achieves a smaller makespan gap than DANIEL in seven of the eight settings across greedy and sampling decoding, while also requiring less inference time.

\begin{table*}[t]
\centering
\caption{Performance comparison on medium- and large-scale deterministic FJSP instances and public benchmarks. Gap (\%) denotes the percentage makespan difference from OR-Tools, and schedule generation time is reported. Lower values are better.}
\label{tab:large}
\begin{adjustbox}{width=\textwidth}
\begin{tabular}{c|c|c|ccc|ccc|ccc|ccc}
\toprule
\multirow{4}{*}{Data} &
\multirow{4}{*}{Size} &
\multirow{2}{*}{OR-Tools} &
\multicolumn{6}{c|}{Greedy} &
\multicolumn{6}{c}{Sampling} \\

\cmidrule(lr){4-9}\cmidrule(l){10-15}

&&&
\multicolumn{3}{c|}{DANIEL} &
\multicolumn{3}{c|}{PLAN} &
\multicolumn{3}{c|}{DANIEL} &
\multicolumn{3}{c}{PLAN} \\

\cmidrule(lr){3-3}
\cmidrule(lr){4-6}
\cmidrule(lr){7-9}
\cmidrule(lr){10-12}
\cmidrule(l){13-15}

&&
Makespan &
Makespan & Gap & Time &
Makespan & Gap & Time &
Makespan & Gap & Time &
Makespan & Gap & Time \\
\midrule

\multirow{4}{*}{SD1}
&30$\times$10
&274.67
&293.48& 6.85&2.44
&\textbf{288.35}&\textbf{4.98}&\textbf{1.91}
&291.66& 6.19&8.69
&\textbf{286.38}& \textbf{4.26}&\textbf{8.09}\\

&40$\times$10
&365.96
&385.97& 5.47&3.25
&\textbf{379.28}&\textbf{3.64}&\textbf{2.52}
&386.59& 5.64&12.60
&\textbf{379.91}&\textbf{3.81}&\textbf{11.89}\\

&100$\times$10
&944.20
&933.61&$-$1.12&7.94
&\textbf{920.84}&\textbf{$-$2.47}&\textbf{6.11}
&963.33&2.03&\textbf{52.72}
&\textbf{951.21}&\textbf{0.74}&59.90\\

&200$\times$5
&1884.70
&1893.65& 0.47&8.03
&\textbf{1880.33}&\textbf{$-$0.23}&\textbf{6.55}
&1991.50&5.67&\textbf{45.86}
&\textbf{1984.71}&\textbf{5.31}&53.80\\

\midrule

\multirow{4}{*}{SD2}
&30$\times$10
&692.26
&803.74& 16.10&2.47
&\textbf{775.64}&\textbf{12.04}&\textbf{1.89}
&766.45&10.72&8.76
&\textbf{734.24}&\textbf{6.06}&\textbf{8.01}\\

&40$\times$10
&998.39
&992.15&$-$0.63&3.24
&\textbf{960.53}&\textbf{$-$3.79}&\textbf{2.56}
&963.77&$-$3.47&12.59
&\textbf{926.98}&\textbf{$-$7.15}&\textbf{11.89}\\

&100$\times$10
&2114.50
&2258.44&6.81&7.85
&\textbf{2216.96}&\textbf{4.85}&\textbf{6.20}
&2303.38&8.93&\textbf{54.21}
&\textbf{2236.00}&\textbf{5.75}&61.76\\

&200$\times$5
&5876.30
&\textbf{5945.83}&\textbf{1.18}&8.05
&5948.38&1.23&\textbf{6.47}
&6990.37&18.96&\textbf{45.76}
&\textbf{6802.20}&\textbf{15.76}&54.30\\

\midrule

\multirow{4}{*}{Public}
&\texttt{Mk}
&174.20
&186.92&7.30&0.80
&\textbf{184.88}&\textbf{6.13}&\textbf{0.58}
&181.64&4.27&2.27
&\textbf{180.00}&\textbf{3.33}&\textbf{1.68}\\

&\texttt{rdata}
&935.80
&1029.37&10.00&0.50
&\textbf{1025.34}&\textbf{9.57}&\textbf{0.33}
&983.60&5.11&1.10
&\textbf{980.42}&\textbf{4.77}&\textbf{0.87}\\

&\texttt{edata}
&1028.93
&1188.39&15.50&0.45
&\textbf{1176.48}&\textbf{14.34}&\textbf{0.29}
&1120.95&8.94&1.10
&\textbf{1118.57}&\textbf{8.71}&\textbf{0.83}\\

&\texttt{vdata}
&919.60
&\textbf{944.78}&\textbf{2.74}&0.45
&947.01&2.98&\textbf{0.28}
&\textbf{924.99}&\textbf{0.59}&1.11
&925.34&0.62&\textbf{0.81}\\

\bottomrule
\end{tabular}
\end{adjustbox}
\end{table*}

\begin{table*}[]
\centering
\caption{Performance comparison on the stochastic FJSP benchmark SD3. SPM-PLAN is evaluated with hidden dimensions of 32 and 8. The best makespan and inference time within each decoding strategy are highlighted in bold. Lower values are better.}
\label{tab:sto}
\begin{adjustbox}{width=\textwidth}
\begin{tabular}{c|c|cc|cc|cc|cc|cc|cc}
\toprule
\multirow{4}{*}{Data} &
\multirow{4}{*}{Size} &
\multicolumn{6}{c|}{Greedy} &
\multicolumn{6}{c}{Sampling} \\

\cmidrule(lr){3-8}\cmidrule(l){9-14}

&&
\multicolumn{2}{c|}{SPM-DAN} &
\multicolumn{2}{c|}{SPM-PLAN (32)} &
\multicolumn{2}{c|}{SPM-PLAN (8)} &
\multicolumn{2}{c|}{SPM-DAN} &
\multicolumn{2}{c|}{SPM-PLAN (32)} &
\multicolumn{2}{c}{SPM-PLAN (8)} \\

\cmidrule(lr){3-4}
\cmidrule(lr){5-6}
\cmidrule(lr){7-8}
\cmidrule(lr){9-10}
\cmidrule(lr){11-12}
\cmidrule(l){13-14}

&&
Makespan & Time &
Makespan & Time &
Makespan & Time &
Makespan & Time &
Makespan & Time &
Makespan & Time\\

\midrule

\multirow{8}{*}{SD3}
&10$\times$5
&718.52&0.85
&712.38&0.78
&\textbf{707.96}&\textbf{0.75}
&675.33&1.56
&673.84&1.43
&\textbf{668.68}&\textbf{0.97}\\

&20$\times$5
&1319.79&1.84
&\textbf{1255.50}&1.78
&1263.76&\textbf{1.70}
&1275.93&4.34
&\textbf{1226.40}&4.34
&1238.38&\textbf{2.95}\\

&15$\times$10
&1085.35&2.60
&\textbf{1080.34}&\textbf{2.28}
&1081.44&2.47
&1032.49&8.69
&\textbf{1027.65}&8.85
&1031.24&\textbf{5.94}\\

&20$\times$10
&\textbf{1287.89}&3.47
&1296.59&\textbf{3.16}
&1296.30&3.36
&\textbf{1254.90}&15.36
&1259.91&14.95
&1258.02&\textbf{9.49}\\

&30$\times$10
&1884.34&5.20
&1858.62&\textbf{4.67}
&\textbf{1839.13}&5.00
&1870.09&32.56
&1853.43&32.13
&\textbf{1819.10}&\textbf{19.92}\\

&40$\times$10
&2423.90&7.03
&2395.19&\textbf{6.42}
&\textbf{2374.47}&6.68
&2434.25&55.95
&2409.49&56.34
&\textbf{2362.94}&\textbf{35.35}\\

&100$\times$10
&5567.87&17.56
&5556.26&\textbf{16.57}
&\textbf{5529.57}&16.78
&5759.71&326.11
&5718.01&337.87
&\textbf{5593.38}&\textbf{201.69}\\

&200$\times$5
&\textbf{10489.71}&17.26
&10734.17&\textbf{16.02}
&10667.28&16.63
&10881.35&302.72
&10905.11&314.58
&\textbf{10700.95}&\textbf{177.09}\\



\bottomrule
\end{tabular}
\end{adjustbox}
\end{table*}

\begin{table}[!ht]
\centering
\caption{SPM-PLAN performance across hidden dimensions on SD3, with model size in kB. The best makespan within each decoding strategy is highlighted in bold.}
\label{tab:hidden_dim}
\footnotesize
\begin{tabular}{c|ccccc}
\toprule
Hidden Dim. & 32 & 16 & 8 & 4 & 2 \\
\midrule
Model Size & 228 & 126 & 97 & 89 & 85 \\
Greedy & 611.0 & 618.2 & \textbf{610.3} & 620.2 & 611.8 \\
Sampling & 586.8 & 588.7 & \textbf{585.0} & 592.6 & 587.8 \\
\bottomrule
\end{tabular}
\end{table}

To evaluate cross-scale generalization, we train PLAN and DANIEL only on $10\times5$ instances and test them on unseen medium- and large-scale instances and public benchmarks. As shown in Table~\ref{tab:large}, PLAN outperforms DANIEL in all but two settings and even surpasses OR-Tools in several cases. Its advantage is maintained across substantially larger and more diverse problem settings, indicating that PLAN generalizes effectively beyond the smallest training scale.

Supplementary Table S2 further shows that PLAN consistently outperforms representative priority dispatching rules (PDRs)~\cite{PDR}, including shortest processing time (SPT) and most work remaining (MWKR), as well as the DRL-based HGNN across diverse benchmark settings. Moreover, PLAN reduces the parameter count by 53.16\% (28,834 to 13,560) and halves the model size from 136kB to 68kB while achieving better scheduling performance, confirming that its accuracy gains do not come at the cost of model complexity or deployment efficiency.

\begin{table*}[t]
\centering
\caption{Performance comparison between HGT and PLAN on the multi-faceted dynamic FJSP benchmark under different dynamic conditions. The best makespan and inference time are highlighted in bold. Lower values are better.}
\label{tab:mfd}
\begin{adjustbox}{width=\textwidth}
\small
\begin{tabular}{c|cc|cc|cc|cc|cc|cc}
\toprule
\multirow{4}{*}{Dataset}
& \multicolumn{4}{c|}{$p=0.4$, $\sigma=1$, $\mu=0.01$}
& \multicolumn{4}{c|}{$p=0.5$, $\sigma=1$, $\mu=0.01$}
& \multicolumn{4}{c}{$p=0.4$, $\sigma=1$, $\mu=0.015$} \\
\cmidrule(lr){2-5}\cmidrule(lr){6-9}\cmidrule(lr){10-13}
& \multicolumn{2}{c|}{HGT}
& \multicolumn{2}{c|}{PLAN}
& \multicolumn{2}{c|}{HGT}
& \multicolumn{2}{c|}{PLAN}
& \multicolumn{2}{c|}{HGT}
& \multicolumn{2}{c}{PLAN} \\
\cmidrule(lr){2-3}\cmidrule(lr){4-5}
\cmidrule(lr){6-7}\cmidrule(lr){8-9}
\cmidrule(lr){10-11}\cmidrule(lr){12-13}
& Makespan & Time
& Makespan & Time
& Makespan & Time
& Makespan & Time
& Makespan & Time
& Makespan & Time \\
\midrule

10$\times$5
&264.20&2.09&\textbf{259.50}&\textbf{1.98}
&302.00&2.20&\textbf{278.70}&\textbf{1.92}
&278.60&2.13&\textbf{275.60}&\textbf{1.93}\\

15$\times$5
&388.90&2.75&\textbf{379.70}&\textbf{2.34}
&382.80&2.67&\textbf{367.00}&\textbf{2.44}
&\textbf{391.60}&2.92&405.80&\textbf{2.52}\\

20$\times$5
&548.30&3.67&\textbf{492.20}&\textbf{3.00}
&\textbf{516.50}&3.59&518.60&\textbf{3.14}
&\textbf{492.80}&3.29&496.20&\textbf{3.10}\\

20$\times$10
&307.00&2.87&\textbf{303.40}&\textbf{2.67}
&321.30&3.14&\textbf{317.80}&\textbf{2.80}
&338.70&3.26&\textbf{326.40}&\textbf{2.91}\\

30$\times$10
&425.20&4.30&\textbf{397.10}&\textbf{3.52}
&416.80&4.49&\textbf{414.40}&\textbf{3.62}
&428.90&4.45&\textbf{422.70}&\textbf{3.52}\\

40$\times$10
&554.10&5.80&\textbf{515.00}&\textbf{4.31}
&547.30&5.88&\textbf{525.50}&\textbf{4.42}
&542.90&5.76&\textbf{542.50}&\textbf{4.52}\\

80$\times$20 
& 596.50 & 21.32 & \textbf{579.80} & \textbf{8.68} 
& 620.35 & 20.63 & \textbf{617.20} & \textbf{8.48}
& 528.50 & 21.31 & \textbf{527.45} & \textbf{9.01} \\

90$\times$30 
& 569.10 & 35.72 & \textbf{566.80} & \textbf{10.57}
& \textbf{658.25} & 34.44 & 658.55 & \textbf{11.17}
& 564.35 & 35.13 & \textbf{558.60} & \textbf{10.70} \\

\bottomrule
\end{tabular}
\end{adjustbox}
\end{table*}

\subsection{Stochastic FJSP}
\begin{figure}[]
\centering
\def \figH{1.25in}
\includegraphics[height=\figH]{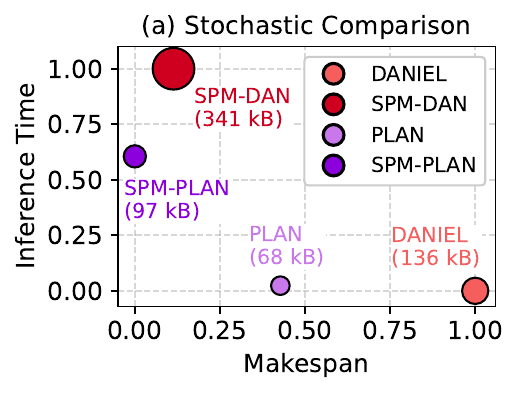}
\hspace{-0.5em}
\includegraphics[height=\figH]{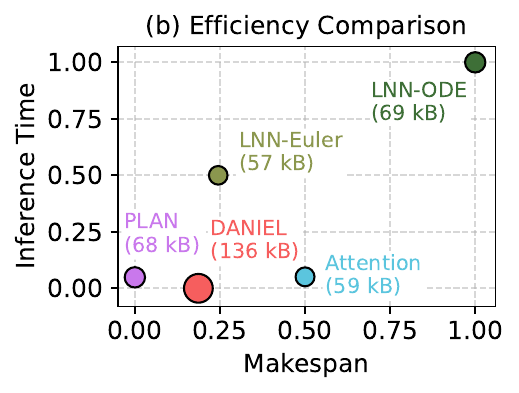}
\caption{Normalized average performance trade-offs across scheduling models. Lower values are better, and marker size indicates model size. (a) Stochastic FJSP. (b) Ablation study.}
\label{fig:pareto}
\end{figure}

\begin{figure}
\centering
\def \figH{1.28in}
\includegraphics[height=\figH]{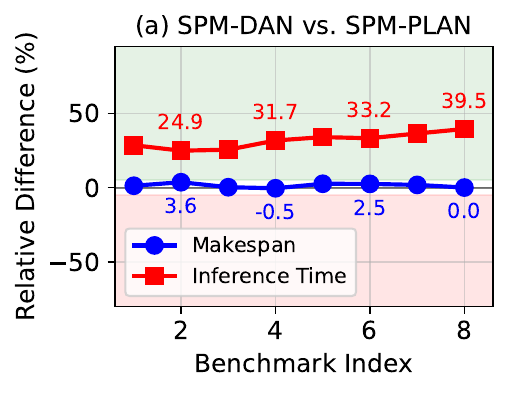}
\hspace{-0.5em}
\includegraphics[height=\figH]{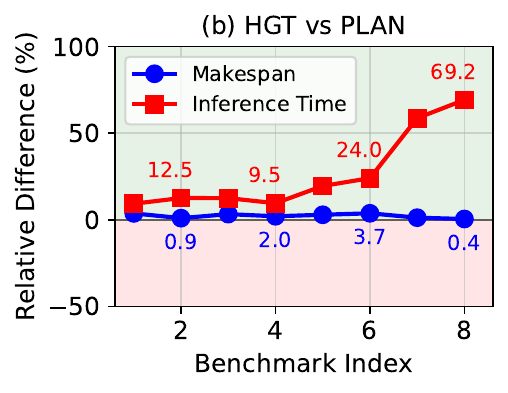}
\caption{Normalized average performance differences across scheduling models. Benchmark indices follow the corresponding setting order in Tables~\ref{tab:sto} and~\ref{tab:mfd}, from smaller to larger problems. Positive values indicate that PLAN performs better. (a) Stochastic FJSP. (b) Multi-faceted dynamic FJSP.}
\label{fig:diff}
\end{figure}

We show the stochastic scheduling results in Table~\ref{tab:sto}, where 32 and 8 denote the SPM hidden dimensions used in the two SPM-PLAN variants, respectively.
Similar to the deterministic FJSP experiments, for medium and large datasets, we evaluate generalization using models trained on 10$\times$5 instances. 
SPM-PLAN achieves the lowest makespan in six of the eight settings under greedy strategy and seven of the eight settings under sampling strategy. For inference time, SPM-PLAN (32) is the fastest in six greedy settings, while SPM-PLAN (8) is the fastest in the remaining two settings and all sampling settings. In particular, SPM-PLAN (8) achieves the lowest makespan in four greedy and five sampling settings, while progressively reducing the parameter count by 78.1\% (77,314 to 16,922) and the model size from 341kB for SPM-DAN to 228kB for SPM-PLAN (32) and 97kB for SPM-PLAN (8).
This observation suggests that PLAN requires only a compact stochastic representation, as its liquid-inspired state updates already capture scheduling-state evolution under processing-time uncertainty. To adapt SPM to PLAN, we tune its hidden dimension on the $10\times5$ training instances. Table~\ref{tab:hidden_dim} further shows that a hidden dimension of 8 achieves the best overall balance between scheduling performance and efficiency.
Figure~\ref{fig:pareto}(a) shows that incorporating SPM improves makespan for both DANIEL and PLAN, although it increases inference time. SPM-DAN therefore performs better than DANIEL under stochastic processing times, while SPM-PLAN similarly improves upon PLAN. Nevertheless, DANIEL and SPM-DAN remain inferior to PLAN and SPM-PLAN, respectively, indicating that PLAN provides a stronger scheduling representation both with and without stochastic modelling.
Figure~\ref{fig:diff}(a) further compares the normalized performance differences across the SD3 benchmark settings. While PLAN maintains generally lower makespan, its inference-time advantage becomes larger as the benchmark index increases, indicating better scalability to larger stochastic FJSP instances.

\subsection{Multi-faceted Dynamic FJSP}
We evaluate PLAN on the multi-faceted dynamic FJSP setting~\cite{MFDFJSP}, which jointly considers dynamic job arrivals, processing-time variation, and machine breakdowns, controlled by $p$, $\sigma$, and $\mu$, respectively. This setting evaluates whether PLAN remains effective under multiple simultaneous scheduling dynamics. Following the original framework, we replace its HGT scheduler with PLAN and retain all other settings unchanged for a fair comparison, such as the hidden dimension of 128, for which both models are larger than those used in the previous experiments. Both models are trained on $10\times5$ instances and evaluated on medium- and large-scale settings to assess cross-scale generalization.

Table~\ref{tab:mfd} shows the results under different dynamic conditions. PLAN achieves lower makespan in most settings and lower inference latency than HGT across all reported settings, while reducing the parameter count by 54.8\% (1,571,099 to 710,939) and the model size from 6.2MB to 2.8MB. Supplementary Tables S6$-$S8 evaluate the two models under additional job arrival rates, machine breakdown rates, and processing time variations, further confirming the robustness of PLAN. These results show that PLAN offers a better performance-efficiency trade-off than a deep attention-based HGT learner.
Similar to Figure \ref{fig:diff}(a), Figure \ref{fig:diff}(b) examines the performance differences across benchmark settings, following the order in Table~\ref{tab:mfd}. The makespan improvement remains relatively stable, whereas the time gap widens at larger problem scales.

\subsection{Ablation Study}
We conduct an ablation study to examine the contributions of the liquid-inspired state update and lightweight attention components in PLAN. We compare four representation learning architectures, including LNN-ODE with continuous liquid dynamics, LNN-Euler with parallel liquid approximation, Attention with lightweight attention only, and the full version of PLAN, with DANIEL as the baseline. The variants are evaluated on representative $10\times5$ and $100\times10$ deterministic FJSP instances. Figure~\ref{fig:pareto}(b) compares their scheduling performance, inference efficiency, and model complexity to illustrate the resulting trade-offs, while Supplementary Table S9 reports the detailed makespan and inference time.
The figure shows that replacing continuous ODE dynamics with the Euler approximation substantially reduces inference time and improves scheduling makespan. LNN-Euler already achieves performance comparable to DANIEL with a much smaller model. In contrast, the lightweight Attention-only model is computationally efficient but exhibits weaker representation capability. By combining lightweight attention with the liquid-inspired state update, PLAN achieves the best overall trade-off between scheduling performance, inference efficiency, and model complexity, while reducing the model size significantly from DANIEL's 136kB to 68kB.

\section{Conclusion}
In this paper, we proposed PLAN, a lightweight representation learning framework that combines parallel liquid-inspired state updates with lightweight attention for DRL-based FJSP. PLAN provides a compact alternative to deep attention-based representation learners by modeling scheduling-state evolution while retaining efficient global context aggregation. On deterministic FJSP benchmarks, PLAN achieves better scheduling performance with lower inference latency and a smaller model size than representative baselines. PLAN also maintains its advantage when trained on small instances and tested on unseen larger settings. In stochastic FJSP, PLAN integrates effectively with scenario aggregation, where the compact SPM-PLAN variant reduces the model size from 341kB to 97kB while outperforming SPM-DAN in most settings. In multi-faceted dynamic FJSP, PLAN improves both makespan and inference latency over HGT. The ablation results further confirm that Euler-based liquid approximation and lightweight attention jointly provide the best performance-efficiency trade-off. Overall, PLAN enables faster scheduling decisions and more scheduling trials within a fixed computational budget.

\bibliography{aaai2027}

@article{kwan2026rela,
  title={ReLA: Representation Learning and Aggregation for Job Scheduling with Reinforcement Learning},
  author={Kwan, Zhengyi and Zhang, Wei and Ng, Aik Beng and Wang, Zhengkui and See, Simon},
  journal={arXiv preprint arXiv:2601.03646},
  year={2026}
}

@article{PDR,
author = {Veronique Sels and Nele Gheysen and Mario Vanhoucke},
title = {A comparison of priority rules for the job shop scheduling problem under different flow time- and tardiness-related objective functions},
journal = {International Journal of Production Research},
volume = {50},
number = {15},
pages = {4255--4270},
year = {2012},
publisher = {Taylor \& Francis},
doi = {10.1080/00207543.2011.611539},
URL = {https://doi.org/10.1080/00207543.2011.611539},
eprint = {https://doi.org/10.1080/00207543.2011.611539}
}

@article{MLD,
author = {Wang, Xinwei and Yu, Xinyong and Wang, Zhaopan and Si, Zeying and Wu, Guangyu and Su, Xi-chao and Qu, Shaohui and Xiong, Biao and Peng, Haijun and Li, Xin and Wang, Lei},
year = {2026},
month = {06},
pages = {},
title = {Machine Learning-Driven Combinatorial Optimization: A Systematic Review},
journal = {Archives of Computational Methods in Engineering},
doi = {10.1007/s11831-026-10679-4}
}

@inproceedings{NCO,
  title={Neural Combinatorial Optimization for Stochastic Flexible Job Shop Scheduling Problems},
  author={Smit, Igor G. and Wu, Yaoxin and Troubil, Pavel and Zhang, Yingqian and Nuijten, Wim P.M.},
  booktitle={Proceedings of the AAAI Conference on Artificial Intelligence},
  volume={39(25)},
  pages={26678--26687},
  year={2025},
  month={4},
  doi={10.1609/aaai.v39i25.34870},
  url={https://ojs.aaai.org/index.php/AAAI/article/view/34870}
}

@INPROCEEDINGS{ADP,
  author={Kim, Min-Soo and Oh, Seog-Chan and Chang, Eun Hyo and Lee, Sangheon and Wells, James W. and Arinez, Jorge and Jang, Young Jae},
  booktitle={2022 IEEE 18th International Conference on Automation Science and Engineering (CASE)}, 
  title={A dynamic programming-based heuristic algorithm for a flexible job shop scheduling problem of a matrix system in automotive industry}, 
  year={2022},
  volume={},
  number={},
  pages={777-782},
  doi={10.1109/CASE49997.2022.9926440}
}

@article{ATA,
title = {An integrated approach for scheduling health care activities in a hospital},
journal = {European Journal of Operational Research},
volume = {264},
number = {2},
pages = {756-773},
year = {2018},
issn = {0377-2217},
doi = {https://doi.org/10.1016/j.ejor.2017.06.051},
url = {https://www.sciencedirect.com/science/article/pii/S0377221717305921},
author = {Robert L. Burdett and Erhan Kozan}
}

@article{FMS,
title = {A simulation optimization framework to solve Stochastic Flexible Job-Shop Scheduling Problems—Case: Semiconductor manufacturing},
journal = {Computers \& Operations Research},
volume = {163},
pages = {106508},
year = {2024},
issn = {0305-0548},
doi = {https://doi.org/10.1016/j.cor.2023.106508},
url = {https://www.sciencedirect.com/science/article/pii/S0305054823003726},
author = {Ensieh Ghaedy-Heidary and Erfan Nejati and Amir Ghasemi and S. Ali Torabi}
}

@article{ROF,
author = {Xie, Jin and Gao, Liang and Peng, Kunkun and Li, Xinyu and Li, Haoran},
title = {Review on flexible job shop scheduling},
journal = {IET Collaborative Intelligent Manufacturing},
volume = {1},
number = {3},
pages = {67-77},
doi = {https://doi.org/10.1049/iet-cim.2018.0009},
url = {https://ietresearch.onlinelibrary.wiley.com/doi/abs/10.1049/iet-cim.2018.0009},
eprint = {https://ietresearch.onlinelibrary.wiley.com/doi/pdf/10.1049/iet-cim.2018.0009},
year = {2019}
}

@article{mkdir,
author = {Brandimarte, Paolo},
title = {Routing and scheduling in a flexible job shop by tabu search},
year = {1993},
issue_date = {1993},
publisher = {J. C. Baltzer AG, Science Publishers},
address = {USA},
volume = {41},
number = {1–4},
issn = {0254-5330},
journal = {Ann. Oper. Res.},
month = may,
pages = {157–183},
numpages = {27}
}

@article{Hurink,
title = {An elitist nondominated sorting hybrid algorithm for multi-objective flexible job-shop scheduling problem with sequence-dependent setups},
journal = {Knowledge-Based Systems},
volume = {173},
pages = {83-112},
year = {2019},
issn = {0950-7051},
doi = {https://doi.org/10.1016/j.knosys.2019.02.027},
url = {https://www.sciencedirect.com/science/article/pii/S0950705119300887},
author = {Z.C. Li and B. Qian and R. Hu and L.L. Chang and J.B. Yang}
}

@article{AEH,
title = {An effective hybrid genetic algorithm and tabu search for flexible job shop scheduling problem},
journal = {International Journal of Production Economics},
volume = {174},
pages = {93-110},
year = {2016},
issn = {0925-5273},
doi = {https://doi.org/10.1016/j.ijpe.2016.01.016},
url = {https://www.sciencedirect.com/science/article/pii/S0925527316000177},
author = {Xinyu Li and Liang Gao}
}

@article{AND,
title = {A neural-driven constructive heuristic for the flexible job shop scheduling problem: An efficient alternative to complex deep learning methods},
journal = {Computers \& Operations Research},
volume = {191},
pages = {107444},
year = {2026},
issn = {0305-0548},
doi = {https://doi.org/10.1016/j.cor.2026.107444},
url = {https://www.sciencedirect.com/science/article/pii/S0305054826000626},
author = {Mariusz Kaleta and Tomasz Śliwiński}
}

@article{FJS,
title = {Flexible Job Shop Scheduling Problem using graph neural networks and reinforcement learning},
journal = {Computers \& Operations Research},
volume = {182},
pages = {107139},
year = {2025},
issn = {0305-0548},
doi = {https://doi.org/10.1016/j.cor.2025.107139},
url = {https://www.sciencedirect.com/science/article/pii/S0305054825001674},
author = {Xi Liu and Xin Chen and Vincent Chau and Jedrzej Musial and Jacek Blazewicz}
}

@ARTICLE{DANIEL,
  author={Wang, Runqing and Wang, Gang and Sun, Jian and Deng, Fang and Chen, Jie},
  journal={IEEE Transactions on Neural Networks and Learning Systems}, 
  title={Flexible Job Shop Scheduling via Dual Attention Network-Based Reinforcement Learning}, 
  year={2024},
  volume={35},
  number={3},
  pages={3091-3102},
  doi={10.1109/TNNLS.2023.3306421}}

@ARTICLE{HGNN,
  author={Song, Wen and Chen, Xinyang and Li, Qiqiang and Cao, Zhiguang},
  journal={IEEE Transactions on Industrial Informatics}, 
  title={Flexible Job-Shop Scheduling via Graph Neural Network and Deep Reinforcement Learning}, 
  year={2023},
  volume={19},
  number={2},
  pages={1600-1610},
  doi={10.1109/TII.2022.3189725}}

@article{LNN,
title = {A novel uncertainty-aware liquid neural network for noise-resilient time series forecasting and classification},
journal = {Chaos, Solitons \& Fractals},
volume = {193},
pages = {116130},
year = {2025},
issn = {0960-0779},
doi = {https://doi.org/10.1016/j.chaos.2025.116130},
url = {https://www.sciencedirect.com/science/article/pii/S0960077925001432},
author = {Muhammed Halil Akpinar and Orhan Atila and Abdulkadir Sengur and Massimo Salvi and U.R. Acharya}
}

@article{ODE,
title = {The continuous memory: A neural network with ordinary differential equations for continuous-time series analysis},
journal = {Applied Soft Computing},
volume = {167},
pages = {112275},
year = {2024},
issn = {1568-4946},
doi = {https://doi.org/10.1016/j.asoc.2024.112275},
url = {https://www.sciencedirect.com/science/article/pii/S1568494624010494},
author = {Bo Li and Haoyu Chen and Zhiyong An and Yuan Yu and Ying Jia and Long Chen and Mingyan Sun}
}

@inproceedings{DLNET,
  title = {When Smaller Wins: Dual-Stage Distillation and Pareto-Guided Compression of Liquid Neural Networks for Edge Battery Prognostics},
  author = {Kannan, Dhivya Dharshini and Li, Wei and Zhang, Wei and Wang, Jianbiao and Seh, Zhi Wei and Ng, Man-Fai},
  publisher = {28th International Conference on Pattern Recognition, ICPR},
  year = {2026}
}

@article{Transformer,
title = {A comprehensive survey on applications of transformers for deep learning tasks},
journal = {Expert Systems with Applications},
volume = {241},
pages = {122666},
year = {2024},
issn = {0957-4174},
doi = {https://doi.org/10.1016/j.eswa.2023.122666},
url = {https://www.sciencedirect.com/science/article/pii/S0957417423031688},
author = {Saidul Islam and Hanae Elmekki and Ahmed Elsebai and Jamal Bentahar and Nagat Drawel and Gaith Rjoub and Witold Pedrycz}
}

@article{MFDFJSP,
title = {Multi-faceted dynamic flexible job shop scheduling via heterogeneous graph transformer and deep reinforcement learning},
journal = {Expert Systems with Applications},
volume = {303},
pages = {130532},
year = {2026},
issn = {0957-4174},
doi = {https://doi.org/10.1016/j.eswa.2025.130532},
url = {https://www.sciencedirect.com/science/article/pii/S0957417425041478},
author = {Guang Liu and Mengqi Liao and Wei Chen and Zhiyu Zhang and Huaiyu Wan and Youfang Lin}
}

\clearpage
\onecolumn
\Large   
\appendix
\section{Supplementary material}
This supplementary material provides additional experimental settings, results, and details. 

\subsection{Deterministic FJSP}

Supplementary Table~\ref{tab:det_settings} provides the complete hyperparameter settings for training via PPO, PLAN, and generating the dataset for deterministic FJSP.

\begin{table}[H]
\centering
\caption{Training, PLAN architecture, and dataset generation settings used for deterministic FJSP experiments.}
\label{tab:det_settings}
\begin{adjustbox}{width=\textwidth}
\small
\begin{tabular}{lll}
\toprule
\textbf{Training (PPO)} & \textbf{PLAN Architecture} & \textbf{Deterministic Dataset} \\
\midrule
Optimizer: Adam & Operation feature dim: 10 & Processing time: $[1,99]$ \\
Learning rate: $3\times10^{-4}$ & Machine feature dim: 8 & Compatible machines/operation: 1 - 5 \\
Discount factor ($\gamma$): 1.0 & Hidden dimensions: [32, 8] & Operations/job: Equal to \# machines \\
PPO epochs: 4 & Attention heads (Operation): [4,4] & Training instances: 100 \\
PPO clip ($\epsilon$): 0.2 & Attention heads (Machine): [4,4] & Training size: $10\times5$ - $20\times10$\\
GAE ($\lambda$): 0.98 & Actor hidden dim: 64 & Test instances: 100 \\
Mini-batch size: 1024 & Critic hidden dim: 64 & Evaluation: Greedy and Sampling (100) \\
Training environments: 20 & Actor/Critic MLP layers: 3 & Generalization: Models trained on $10\times5$ \\
Maximum updates: 1000 & Dropout: 0 & Test sizes: $10\times5$ - $200\times5$ \\
Validation interval: 10 & Model size: 68 kB & \\
Training/Test seed: 0-4 & & \\
\bottomrule
\end{tabular}
\end{adjustbox}
\end{table}


In Supplementary Table~\ref{tab:comparison}, we compare our proposed PLAN-based scheduling with existing solutions like OR-Tools, top Priority dispatching rules (SPT, MWKR), and DRL-based methods (HGNN, DANIEL). The results show that PLAN consistently achieves superior scheduling quality, outperforming both conventional optimization and state-of-the-art learning-based baselines.

\begin{table}[H]
\centering
\caption{Comparison with OR-Tools, PDRs, and DRL-based methods on benchmark instances.}
\label{tab:comparison}
\begin{adjustbox}{width=\textwidth}
\begin{tabular}{ll|c|cc|ccc|ccc}
\toprule

\multirow{3}{*}{Data} &
\multirow{3}{*}{Size} &
\multirow{3}{*}{OR-Tools} &
\multicolumn{2}{c|}{PDRs} &
\multicolumn{3}{c|}{Greedy (DRL)} &
\multicolumn{3}{c}{Sampling (DRL)} \\

\cmidrule(lr){4-5}
\cmidrule(lr){6-8}
\cmidrule(l){9-11}

&&&
SPT &
MWKR &
HGNN &
DANIEL &
PLAN &
HGNN &
DANIEL &
PLAN \\

\midrule

\multirow{6}{*}{SD1}
&10$\times$5
&96.32
&129.82
&113.23
&111.67
&107.97
&\textbf{107.36}
&105.59
&102.36
&\textbf{101.48}\\

&20$\times$5
&188.15
&230.48
&209.78
&211.22
&\textbf{197.66}
&197.78
&207.53
&193.83
&\textbf{193.14}\\

&15$\times$10
&143.53
&198.33
&171.25
&166.92
&160.78
&\textbf{159.21}
&160.86
&152.51
&\textbf{150.68}\\

&20$\times$10
&195.98
&255.17
&216.11
&215.78
&199.14
&\textbf{198.21}
&214.81
&195.30
&\textbf{193.38}\\

&30$\times$10
&274.67
&350.07
&312.93
&314.71
&293.48
&\textbf{288.35}
&308.55
&291.66
&\textbf{286.38}\\

&40$\times$10
&365.96
&445.17
&414.82
&417.87
&385.97
&\textbf{379.28}
&410.76
&386.59
&\textbf{379.91}\\

\midrule

\multirow{6}{*}{SD2}
&10$\times$5
&326.24
&514.39
&549.28
&553.61
&413.91
&\textbf{407.86}
&483.90
&366.59
&\textbf{362.46}\\

&20$\times$5
&602.04
&835.94
&1026.03
&1059.04
&673.28
&\textbf{661.45}
&962.90
&632.85
&\textbf{624.68}\\

&15$\times$10
&377.17
&703.07
&830.53
&807.47
&588.14
&\textbf{587.84}
&756.07
&519.78
&\textbf{515.61}\\

&20$\times$10
&464.16
&829.14
&1040.69
&1045.82
&606.14
&\textbf{603.43}
&990.37
&552.09
&\textbf{550.11}\\

&30$\times$10
&692.26
&1105.99
&1539.67
&1564.57
&803.74
&\textbf{775.64}
&1486.56
&766.45
&\textbf{734.24}\\

&40$\times$10
&998.39
&1357.16
&2037.65
&2048.96
&992.15
&\textbf{960.53}
&1976.25
&963.77
&\textbf{926.98}\\

\bottomrule
\end{tabular}
\end{adjustbox}
\end{table}

\subsection{Stochastic FSJP}
Supplementary Table~\ref{tab:stoc_settings} provides the complete hyperparameter settings for training via PPO, SPM-PLAN, and generating the dataset for stochastic FJSP.

\begin{table}[H]
\centering
\caption{Training, SPM-PLAN architecture, and stochastic dataset settings used for SFJSP experiments.}
\label{tab:stoc_settings}
\begin{adjustbox}{width=\textwidth}
\small
\begin{tabular}{lll}
\toprule
\textbf{Training (PPO)} & \textbf{SPM-PLAN Architecture} & \textbf{Stochastic Dataset} \\
\midrule
Optimizer: Adam & Operation feature dim: 10 & Processing time: $[1,99]$ \\
Learning rate: $3\times10^{-4}$ & Machine feature dim: 8 & Compatible machines/operation: 1 - 5 \\
Discount factor ($\gamma$): 1.0 & Hidden dimensions: [32, 8] & Operations/job: Equal to \# machines \\
PPO epochs: 4 & Attention heads (Operation): [4,4] & Training instances: 100 \\
PPO clip ($\epsilon$): 0.2 & Attention heads (Machine): [4,4] & Training size: $10\times5$ - $20\times10$ \\
GAE ($\lambda$): 0.98 & SAA Attention: Enabled & Test instances: 100 \\
Mini-batch size: 143 & SAA attention dim: 8 & Input realizations: 100 \\
Gradient accumulation: 7 & Scenario aggregation: Mean & Evaluation realizations: 1000 \\
Training environments: 20 & Actor/Critic hidden dim: 64 & Variance: Random (Lognormal) \\
Maximum updates: 1000 & Actor/Critic MLP layers: 3 & Training method: SAA \\
Validation interval: 10 & Dropout: 0 & Objective: VaR ($\alpha=0.95$) \\
Training seed: 400 & & Generalization: Models trained on $10\times5$ \\
Test seed: 50 & & Test sizes: $10\times5$ - $200\times5$ \\
\bottomrule
\end{tabular}
\end{adjustbox}
\end{table}







\begin{figure}[h]
    \centering
    \includegraphics[width=1\linewidth]{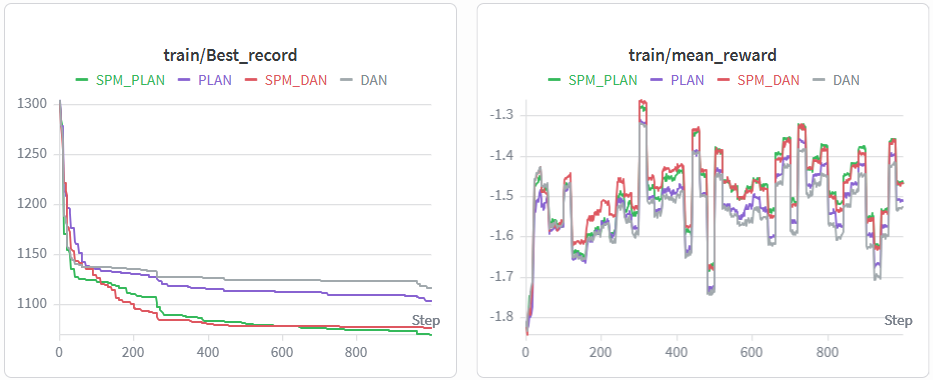}
    \caption{Stochastic FJSP (SD3) training behavior}
    \label{fig:sd3_train}
\end{figure}

Supplementary Fig.~\ref{fig:sd3_train} shows the training behaviour of PLAN and DANIEL with and without SPM modules under the stochastic scenario. PLAN without SPM achieves significantly better reward and attains the best-record makespan throughout training than DAN. With SPM, SPM-PLAN converges to a higher final reward than SPM-DAN, demonstrating its effectiveness in stochastic aggregation settings.

Table~\ref{tab:stochastic1} shows the makespan and inference time comparison of PLAN and DANIEL with and without SPM, across small, medium, and large instances. Without SPM, PLAN shows consistently higher scheduling makespan than DAN. SPM improves scheduling quality of the FJSP with stochastic processing times, with SPM-PLAN achieving performance gains and reducing the inference time by more than 50\% compared to SPM-DAN, offering an efficient solution.

\begin{table}[H]
\centering
\caption{Performance ${VaR}_{\alpha}(C_{\max})$ and Efficiency comparison on stochastic FJSP}
\label{tab:stochastic1}
\begin{adjustbox}{width=\textwidth}
\begin{tabular}{lll|cc|cc|cc|cc|cc}
\toprule
\multirow{2}{*}{Data} &
\multirow{2}{*}{Decoding} &
\multirow{2}{*}{Size} &
\multicolumn{2}{c|}{DAN} &
\multicolumn{2}{c|}{SPM-DAN} &
\multicolumn{2}{c|}{PLAN} &
\multicolumn{2}{c|}{SPM-PLAN (32)} &
\multicolumn{2}{c}{SPM-PLAN (8)}\\

\cmidrule(lr){4-5}
\cmidrule(lr){6-7}
\cmidrule(lr){8-9}
\cmidrule(lr){10-11}
\cmidrule(l){12-13}

&&
  &
Makespan & Time &
Makespan & Time &
Makespan & Time &
Makespan & Time &
Makespan & Time\\

\midrule

\multirow{16}{*}{SD3}

&\multirow{8}{*}{Greedy}
&10$\times$5
&733.92&1.07
&718.52&0.85
&740.96&0.88
&712.38&0.78
&\textbf{707.96}&\textbf{0.75}\\

&&20$\times$5
&1341.80&2.05
&1319.79&1.84
&1297.29&1.66
&\textbf{1255.50}&1.78
&1263.76&\textbf{1.70}\\

&&15$\times$10
&1127.60&3.19
&1085.35&2.60
&1114.99&2.42
&\textbf{1080.34}&\textbf{2.28}
&1081.44&2.47\\

&&20$\times$10
&1376.29&4.24
&\textbf{1287.89}&3.47
&1362.11&3.57
&1296.59&\textbf{3.16}
&1296.30&3.36\\

&&30$\times$10
&1971.57&6.18
&1884.34&5.20
&1970.33&5.20
&1858.62&\textbf{4.67}
&\textbf{1839.13}&5.00\\

&&40$\times$10
&2541.06&8.51
&2423.90&7.03
&2538.39&6.93
&2395.19&\textbf{6.42}
&\textbf{2374.47}&6.68\\

&&100$\times$10
&6019.35&20.93
&5567.87&17.56
&5795.76&17.89
&5556.26&\textbf{16.57}
&\textbf{5529.57}&16.78\\

&&200$\times$5
&11531.42&21.18
&\textbf{10489.71}&17.26
&10633.82&16.95
&10734.17&\textbf{16.02}
&10667.28&16.63\\

\cmidrule(lr){2-13}

&\multirow{8}{*}{Sampling}
&10$\times$5
&684.56&1.24
&675.33&1.56
&685.87&1.12
&673.84&1.43
&\textbf{668.68}&\textbf{0.97}\\

&&20$\times$5
&1291.56&2.45
&1275.93&4.34
&1262.44&2.32
&\textbf{1226.40}&4.34
&1238.38&\textbf{2.95}\\

&&15$\times$10
&1076.24&4.38
&1032.49&8.69
&1065.77&4.01
&\textbf{1027.65}&8.85
&1031.24&\textbf{5.94}\\

&&20$\times$10
&1340.44&6.60
&\textbf{1254.90}&15.36
&1324.40&6.23
&1259.91&14.95
&1258.02&\textbf{9.49}\\

&&30$\times$10
&1933.25&11.39
&1870.09&32.56
&1943.70&11.38
&1853.43&32.13
&\textbf{1819.10}&\textbf{19.92}\\

&&40$\times$10
&2522.82&18.42
&2434.25&55.95
&2530.06&18.01
&2409.49&56.34
&\textbf{2362.94}&\textbf{35.35}\\

&&100$\times$10
&6088.92&85.62
&5759.71&326.11
&5886.58&94.76
&5718.01&337.87
&\textbf{5593.38}&\textbf{201.69}\\

&&200$\times$5
&11547.33&64.40
&10881.35&302.72
&11102.79&70.76
&10905.11&314.58
&\textbf{10700.95}&\textbf{177.09}\\

\midrule

\multicolumn{3}{l|}{Model Size (kB)}
&136&&341&&68&&228&&97\\

\bottomrule
\end{tabular}
\end{adjustbox}
\end{table}

\subsection{Multi-faceted Dynamic FJSP}
Supplementary Table~\ref{tab:mfdfjsp_settings} provides the complete hyperparameter settings for training, PLAN, and generating the dataset for Multi-faceted Dynamic FJSP.

\begin{table}[H]
\centering
\caption{Training, PLAN architecture, and MFDFJSP dataset settings.}
\label{tab:mfdfjsp_settings}
\begin{adjustbox}{width=\textwidth}
\small
\begin{tabular}{lll}
\toprule
\textbf{Training {PPO}} & \textbf{PLAN Architecture} & \textbf{Multi-faceted Dynamic Dataset} \\
\midrule
Optimizer = AdamW &
Embedding dimension = 128 &
Dynamic job arrival ($p$): 0.3, 0.4, 0.5 \\

Learning rate = $5\times10^{-5}$ &
Hidden dimension = 256 &
Processing time variability ($\sigma$): 0.5, 1, 2 \\

Weight decay = $10^{-2}$ &
Encoder layers = 4 &
Machine failure rate ($\mu$): 0.0005, 0.001, 0.0015 \\

Batch size = 64 &
Attention heads = 8 &
Training size = $10\times5$ \\

Training iterations = 100 &
Key dimension = 16 &
Test sizes = $10\times5$--$90\times30$ \\

Epochs / iteration = 3 &
Dropout = 0.2 &
Maximum operations = 400 \\

Gradient clipping = 1.0 &
Operation feature dim = 7 &
Number of cases = 10 \\

PPO clip = 0.15 &
Machine feature dim = 4 &
Generalization: trained on $10\times5$\\

Discount factors = 0.925 &
Arc feature dim = 2 &
Framework: HGAN replaced by PLAN \\

Entropy coefficient = 0.01 &
Representation learner = PLAN &
\\

Policy / Value loss = 1.0 / 1.0 &
&
\\
\bottomrule
\end{tabular}
\end{adjustbox}
\end{table}

For Multi-faceted dynamic FJSP, we consider different dynamic settings by varying these parameters: dynamic job arrivals ($p$), variable processing times ($\sigma$), and machine breakdowns ($\mu$). Table~\ref{tab:dynamic_degree} presents the results under different dynamic degrees of 0.3, 0.4, and 0.5 with $\sigma=1$, $\mu=0.01$. Table~\ref{tab:failure_rate} shows the results under different rates of machine failure (0.005, 0.01, and 0.015) with $p=0.4$, $\sigma=1$. Table~\ref{tab:stochasticity} shows the results under different processing time probabilities, $\sigma=0.5, 1, 2$ with $p=0.4$, $\mu=0.01$. Across most settings, PLAN consistently outperforms HGT, achieving lower makespan and faster inference

\begin{table}[H]
\centering
\caption{Performance comparison under different dynamic degrees.}
\label{tab:dynamic_degree}
\begin{adjustbox}{width=\textwidth}
\small
\setlength{\tabcolsep}{3pt}
\begin{tabular}{l|cc|cc|cc|cc|cc|cc}
\toprule
\multirow{3}{*}{Dataset}
& \multicolumn{4}{c|}{$p=0.3$, $\sigma=1$, $\mu=0.01$}
& \multicolumn{4}{c|}{$p=0.4$, $\sigma=1$, $\mu=0.01$}
& \multicolumn{4}{c}{$p=0.5$, $\sigma=1$, $\mu=0.01$} \\
\cmidrule(lr){2-5}\cmidrule(lr){6-9}\cmidrule(lr){10-13}
& \multicolumn{2}{c|}{HGT}
& \multicolumn{2}{c|}{PLAN}
& \multicolumn{2}{c|}{HGT}
& \multicolumn{2}{c|}{PLAN}
& \multicolumn{2}{c|}{HGT}
& \multicolumn{2}{c}{PLAN} \\
\cmidrule(lr){2-3}\cmidrule(lr){4-5}
\cmidrule(lr){6-7}\cmidrule(lr){8-9}
\cmidrule(lr){10-11}\cmidrule(lr){12-13}
& Makespan & Time (s)
& Makespan & Time (s)
& Makespan & Time (s)
& Makespan & Time (s)
& Makespan & Time (s)
& Makespan & Time (s) \\
\midrule

10$\times$5
&266.00&2.16&\textbf{264.40}&\textbf{1.88}
&264.20&2.09&\textbf{259.50}&\textbf{1.98}
&302.00&2.20&\textbf{278.70}&\textbf{1.92}\\

15$\times$5
&405.60&2.81&\textbf{368.10}&\textbf{2.41}
&388.90&2.75&\textbf{379.70}&\textbf{2.34}
&382.80&2.67&\textbf{367.00}&\textbf{2.44}\\

20$\times$5
&509.10&3.35&\textbf{471.00}&\textbf{2.93}
&548.30&3.67&\textbf{492.20}&\textbf{3.00}
&\textbf{516.50}&3.59&518.60&\textbf{3.14}\\

20$\times$10
&314.00&2.93&\textbf{301.60}&\textbf{2.63}
&307.00&2.87&\textbf{303.40}&\textbf{2.67}
&321.30&3.14&\textbf{317.80}&\textbf{2.80}\\

30$\times$10
&392.60&4.10&\textbf{386.10}&\textbf{3.30}
&425.20&4.30&\textbf{397.10}&\textbf{3.52}
&416.80&4.49&\textbf{414.40}&\textbf{3.62}\\

40$\times$10
&524.80&5.57&\textbf{506.60}&\textbf{4.25}
&554.10&5.80&\textbf{515.00}&\textbf{4.31}
&547.30&5.88&\textbf{525.50}&\textbf{4.42}\\

\bottomrule
\end{tabular}
\end{adjustbox}
\end{table}

\begin{table}[H]
\centering
\caption{Performance comparison under different machine failure rates.}
\label{tab:failure_rate}
\begin{adjustbox}{width=\textwidth}
\small
\setlength{\tabcolsep}{3pt}
\begin{tabular}{l|cc|cc|cc|cc|cc|cc}
\toprule
\multirow{3}{*}{Dataset}
& \multicolumn{4}{c|}{$p=0.4$, $\sigma=1$, $\mu=0.005$}
& \multicolumn{4}{c|}{$p=0.4$, $\sigma=1$, $\mu=0.01$}
& \multicolumn{4}{c}{$p=0.4$, $\sigma=1$, $\mu=0.015$} \\
\cmidrule(lr){2-5}\cmidrule(lr){6-9}\cmidrule(lr){10-13}
& \multicolumn{2}{c|}{HGT}
& \multicolumn{2}{c|}{PLAN}
& \multicolumn{2}{c|}{HGT}
& \multicolumn{2}{c|}{PLAN}
& \multicolumn{2}{c|}{HGT}
& \multicolumn{2}{c}{PLAN} \\
\cmidrule(lr){2-3}\cmidrule(lr){4-5}
\cmidrule(lr){6-7}\cmidrule(lr){8-9}
\cmidrule(lr){10-11}\cmidrule(lr){12-13}
& Makespan & Time (s)
& Makespan & Time (s)
& Makespan & Time (s)
& Makespan & Time (s)
& Makespan & Time (s)
& Makespan & Time (s) \\
\midrule

10$\times$5
&259.70&2.02&\textbf{257.50}&\textbf{1.82}
&264.20&2.09&\textbf{259.50}&\textbf{1.98}
&278.60&2.13&\textbf{275.60}&\textbf{1.93}\\

15$\times$5
&384.60&2.62&\textbf{372.40}&\textbf{2.48}
&388.90&2.75&\textbf{379.70}&\textbf{2.34}
&\textbf{391.60}&2.92&405.80&\textbf{2.52}\\

20$\times$5
&484.50&3.26&\textbf{474.30}&\textbf{2.98}
&548.30&3.67&\textbf{492.20}&\textbf{3.00}
&\textbf{492.80}&3.29&496.20&\textbf{3.10}\\

20$\times$10
&296.80&2.90&\textbf{288.30}&\textbf{2.61}
&307.00&2.87&\textbf{303.40}&\textbf{2.67}
&338.70&3.26&\textbf{326.40}&\textbf{2.91}\\

30$\times$10
&403.70&4.11&424.20&\textbf{3.49}
&425.20&4.30&\textbf{397.10}&\textbf{3.52}
&428.90&4.45&\textbf{422.70}&\textbf{3.52}\\

40$\times$10
&524.30&5.62&532.60&\textbf{4.48}
&554.10&5.80&\textbf{515.00}&\textbf{4.31}
&542.90&5.76&\textbf{542.50}&\textbf{4.52}\\

\bottomrule
\end{tabular}
\end{adjustbox}
\end{table}

\begin{table}[H]
\centering
\caption{Performance comparison under different probabilities of processing time.}
\label{tab:stochasticity}
\begin{adjustbox}{width=\textwidth}
\small
\setlength{\tabcolsep}{3pt}
\begin{tabular}{l|cc|cc|cc|cc|cc|cc}
\toprule
\multirow{3}{*}{Dataset}
& \multicolumn{4}{c|}{$p=0.4$, $\sigma=0.5$, $\mu=0.01$}
& \multicolumn{4}{c|}{$p=0.4$, $\sigma=1$, $\mu=0.01$}
& \multicolumn{4}{c}{$p=0.4$, $\sigma=2$, $\mu=0.01$} \\
\cmidrule(lr){2-5}\cmidrule(lr){6-9}\cmidrule(lr){10-13}
& \multicolumn{2}{c|}{HGT}
& \multicolumn{2}{c|}{PLAN}
& \multicolumn{2}{c|}{HGT}
& \multicolumn{2}{c|}{PLAN}
& \multicolumn{2}{c|}{HGT}
& \multicolumn{2}{c}{PLAN} \\
\cmidrule(lr){2-3}\cmidrule(lr){4-5}
\cmidrule(lr){6-7}\cmidrule(lr){8-9}
\cmidrule(lr){10-11}\cmidrule(lr){12-13}
& Makespan & Time (s)
& Makespan & Time (s)
& Makespan & Time (s)
& Makespan & Time (s)
& Makespan & Time (s)
& Makespan & Time (s) \\
\midrule

10$\times$5
&277.80&2.02&\textbf{253.20}&\textbf{1.75}
&264.20&2.09&\textbf{259.50}&\textbf{1.98}
&293.50&2.23&299.30&\textbf{1.98}\\

15$\times$5
&397.50&2.73&\textbf{372.40}&\textbf{2.41}
&388.90&2.75&\textbf{379.70}&\textbf{2.34}
&404.60&2.83&\textbf{402.30}&\textbf{2.47}\\

20$\times$5
&545.60&3.54&\textbf{494.10}&\textbf{3.04}
&548.30&3.67&\textbf{492.20}&\textbf{3.00}
&\textbf{470.20}&3.34&475.60&\textbf{2.90}\\

20$\times$10
&296.30&3.08&\textbf{295.30}&\textbf{2.64}
&307.00&2.87&\textbf{303.40}&\textbf{2.67}
&326.80&2.99&\textbf{305.30}&\textbf{2.63}\\

30$\times$10
&427.10&4.30&\textbf{390.90}&\textbf{3.36}
&425.20&4.30&\textbf{397.10}&\textbf{3.52}
&420.30&4.38&426.60&\textbf{3.52}\\

40$\times$10
&559.60&5.72&\textbf{520.20}&\textbf{4.39}
&554.10&5.80&\textbf{515.00}&\textbf{4.31}
&\textbf{510.30}&5.44&527.90&\textbf{4.52}\\

\bottomrule
\end{tabular}
\end{adjustbox}
\end{table}

\subsection{Detailed Ablation results}

\begin{figure}[H]
    \centering
    \includegraphics[width=0.95\linewidth]{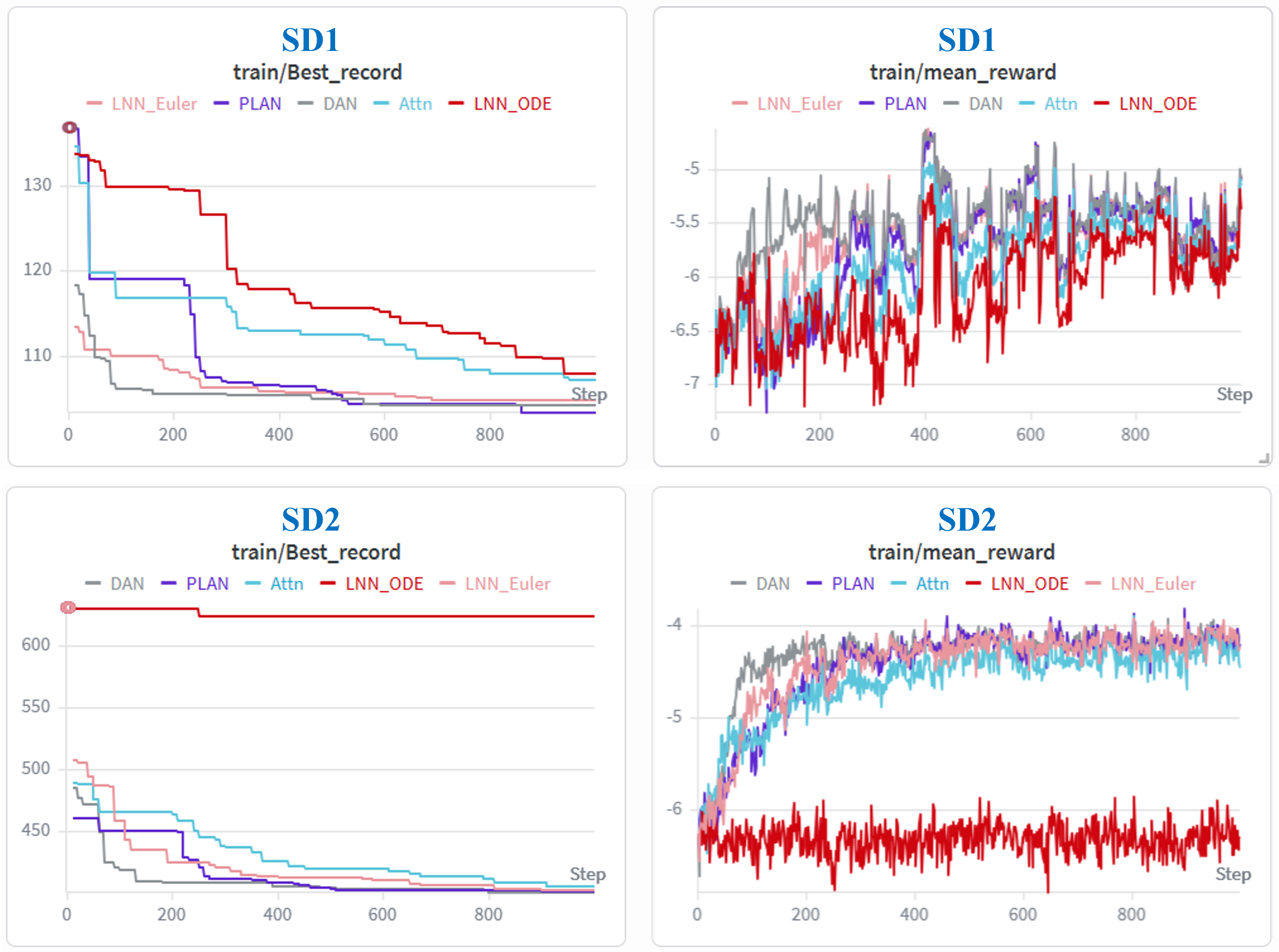}
    \caption{Ablation Training behavior}
    \label{fig:abl_train}
\end{figure}

Supplementary Fig.~\ref{fig:abl_train} shows the training behaviour of LNN with ODE, discretized LNN, Simple Attention, and PLAN compared with DANIEL. The original continuous ODE-based LNN fails to adapt to the discrete FSJP dynamics, whereas discretization with Euler yields performance competitive with DANIEL. By combining the liquid-inspired state update with simple attention, PLAN reaches a higher final training reward.

Supplementary Table~\ref{tab:ablation} shows the makespan and inference time across 10$\times$5 and 100$\times$10 instance sizes in greedy and sampling encoding. To evaluate generalization, the 100$\times$10 instances are tested using the model trained on the 10$\times$5 dataset. The results show that PLAN offers a better performance-efficiency trade-off than DANIEL in both small and large problem scales.

\begin{table}[H]
\centering
\caption{Ablation study on 10x5 and 100x10 dataset}
\label{tab:ablation}
\begin{adjustbox}{width=\textwidth}
\begin{tabular}{lll|ccccc|ccccc}
\toprule
\multirow{2}{*}{Data} &
\multirow{2}{*}{Size} &
\multirow{2}{*}{Metric} &
\multicolumn{5}{c|}{Greedy} &
\multicolumn{5}{c}{Sampling} \\
\cmidrule(lr){4-8} \cmidrule(l){9-13}
&&&
DANIEL & LNN\_ODE & LNN\_Euler & Attention & PLAN &
DANIEL & LNN\_ODE & LNN\_Euler & Attention & PLAN \\
\midrule

\multirow{4}{*}{SD1}
& \multirow{2}{*}{10$\times$5}
& Makespan
& 108.30 & 110.53 & 108.36 & 110.06 & \textbf{107.22}
& 102.37 & 103.43 & 102.19 & 103.57 & \textbf{101.67} \\

&& Inf Time (s)
& 0.48 & 6.70 & 0.49 & 0.31 & 0.33
& 1.15 & 7.90 & 1.19 & 0.82 & 0.81 \\

\cmidrule(lr){2-13}

& \multirow{2}{*}{100$\times$10}
& Makespan
& 933.61 & 1066.36 & 989.55 & 949.65 & \textbf{920.84}
& 963.33 & 1050.78 & 1017.18 & 1007.15 & \textbf{951.21} \\

&& Inf Time (s)
& 7.94 & 154.11 & 43.38 & 6.04 & 6.11
& 52.72 & 435.37 & 92.66 & 60.48 & 59.90 \\

\midrule

\multirow{4}{*}{SD2}
& \multirow{2}{*}{10$\times$5}
& Makespan
& 409.75 & 502.27 & \textbf{407.78} & 415.00 & 407.86
& 363.96 & 421.74 & 362.32 & \textbf{362.12} & 362.46 \\

&& Inf Time (s)
& 0.40 & 6.06 & 0.61 & 0.44 & 0.31
& 1.11 & 7.24 & 1.16 & 0.85 & 0.81 \\

\cmidrule(lr){2-13}

& \multirow{2}{*}{100$\times$10}
& Makespan
& 2258.44 & Invalid & 2232.14 & 2275.00 & \textbf{2216.96}
& 2303.38 & Invalid & 2267.55 & 2459.75 & \textbf{2236.00} \\

&& Inf Time (s)
& 7.94 & -- & 46.80 & 5.96 & 6.20
& 54.21 & -- & 95.99 & 61.80 & 61.76 \\

\midrule

\multicolumn{3}{l|}{Model Size (kB)}
& 135 & 69 & 57 & 59 & 68
& 135 & 69 & 57 & 59 & 68\\

\bottomrule
\end{tabular}
\end{adjustbox}
\end{table}

\end{document}